# Optimal Pattern synthesis of linear antenna array using Ant Hill Colonization Optimization algorithm(AHCOA)

Sunit Shantanu Digamber Fulari[1][0000-0001-8121-2117]

[1] Electronics and Communication dept., Chandigarh University, Mohali, India

**Abstract.** The aim of this paper is to introduce AHCOA to the electromagnetic and antenna community. AHCOA is a new nature inspired meta heuristic algorithm inspired by how there is a hierarchy and departments in the ant hill colonization. It has high probabilistic potential in solving not only unconstrained but also constrained optimization problems. In this paper the AHCOA is applied to linear antenna array for better pattern synthesis in the following ways : By uniform excitation considering equal spacing of the antenna elements with respect to the uniform array. AHCOA is used in obtaining an array pattern to achieve minimum side lobe levels. The results are compared to other state of the art nature based algorithms such as ant lion optimizer, which show a considerable improvement in AHCOA.

## 1  Introduction

The ant species work in a very distinct and efficient manner. They leave behind a trail in their path. This is done by submitting a substance known as pheromones in their path which is left behind to establish kind of connection for the preceding ants. This helps in reestablishing the connection with the next ants to follow the path. Shortest path algorithms is followed in this case also, this makes the final ants follow the pheromonal substance path, mostly the ants follow the ant which reached the first, they kind of have a sense in joining with the shortest path trail. They establish the pheromonal path which leads to their nest, food collection location, or previously left path. In this there is a discussion on linear antenna array which is optimized by the ant colony optimization algorithm.

**II Related work on nature based optimization.**

Nature inspired algorithms are used in providing solutions to engineering problems, they have found uses in applications such as data mining, breast cancer diagnosis, spring design by using the tensions and spring constant variation, pressure vessel design in which the stress and volume is judged to benchmark minimize these parameters(cost),  speed reducer function in which the diameter of the shaft, gear, teeth are set to minimize the cost of the overall design, further the reliability based applications such as solutions to inverse problems and parameter identification, secondly in image processing, feature selection by cuckoo search was effective thirdly travelling salesman problem by discrete cuckoo search algorithm to solve various benchmark instances was implemented(this involves shortest path between cities which has to be implemented using global positioning system), fourthly in vehicle routing is a class related to advances in travelling salesman problem, which involves parameters such as traffic routing, logistic requirements, and the time period of delivery of the consignment, fourthly about flow shop scheduling which is solved either by firefly algorithm, genetic algorithm, ant colony optimization and simulated annealing, fifthly in software testing by Srivastava et.al has proposed discrete firefly algorithm and the cuckoo search algorithm, sixthly in deep belief networks, seventy in swarm robots, next finally in various other optimization where these nature based optimizations are used.

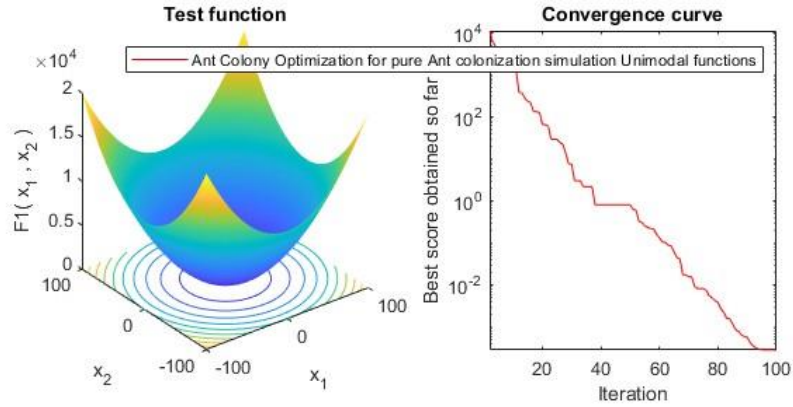

Figure I: Ant colony optimization for pure unimodal functions defined as below.

Figure II: The ant colony optimization for constrained equation of function(Unimodal equations) These are the seven benchmark functions

| Function | Dim | Range | $f_{min}$ |
|---|---|---|---|
| $F_1(x) = \sum_{i=1}^{n} x_i^2$ | 30, 200 | $[-100, 100]$ | 0 |
| $F_2(x) = \sum_{i=1}^{n} |x_i| + \prod_{i=1}^{n} |x_i|$ | 30, 200 | $[-10, 10]$ | 0 |
| $F_3(x) = \sum_{i=1}^{n} \left(\sum_{j=1}^{i} x_j\right)^2$ | 30, 200 | $[-100, 100]$ | 0 |
| $F_4(x) = \max_i \{|x_i|, 1 \leq i \leq n\}$ | 30, 200 | $[-100, 100]$ | 0 |
| $F_5(x) = \sum_{i=1}^{n-1} [100(x_{i+1} - x_i^2)^2 + (x_i - 1)^2]$ | 30, 200 | $[-30, 30]$ | 0 |
| $F_6(x) = \sum_{i=1}^{n} (|x_i + 0.5|)^2$ | 30, 200 | $[-100, 100]$ | 0 |
| $F_7(x) = \sum_{i=1}^{n} i x_i^4 + random[0, 1)$ | 30, 200 | $[-1.28, 1.28]$ | 0 |

When the function is varied from parameters 2 to 3 we obtain.
Figure III: Varied optimization for values from 2 to 3.

### III Proposed work

We in our paper propose an antenna synthesis using AHCOA algorithm, this is used to modify the radiation characteristics of the antenna. The ants build their ant hill systematically with each department playing their role expertly in matters of architecture, food collection, breeding and protection. Our proposed work has shown better results compared to ant lion optimizer in antenna side lobe reduction and the genetic algorithm(GABONST).

Figure III:Non Convergence variation of unimodal function

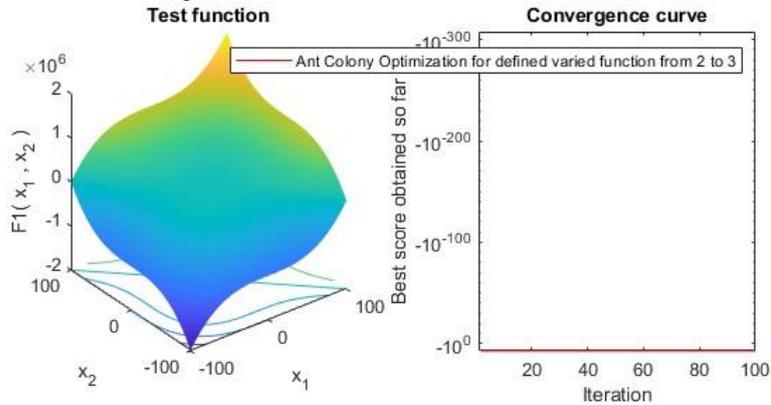

The convergence curve finally deprecated for small variation in the first $F_1(x)$ function value from 2 to 3. The convergence curve has showing faltered values in this case. This shows that ant hill construction not always converges but may increase the rate.

Figure IV: Linear array antenna arrangement.

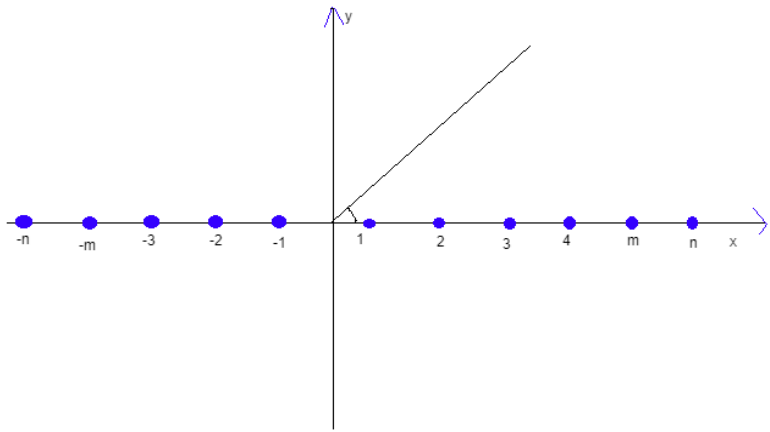

In array designing one of the most important parameter to design is antenna sidelobe, this is most prominent parameter to train in large arrays. Large arrays are dealing with the main lobe, concerning with the directivity of the antenna, where the maximum electromagnetic radiation of the antenna is expelled for interconnection. The radiation characteristics in unwanted directions or the sidelobes which they are known as should be as minimum as possible. This unwanted sidelobe or radiation in minor or side lobes should lobe minimized or reduced. To achieve these characteristics or aim, there is a need to have an equally spaced array i.e. antenna elements which are separated in linear distance equally but with a specific element frequency distribution. Nonuniform spacing of arrays is also possible as another application or a different setup. This setup provides with a low sidelobe at some levels or at some regions. The orientation of the feeding network makes the position of the uniform linear array advantageous. This algorithm helps in solving complex problems which otherwise cannot be used by applying analytical solutions. AHCOA is a new idea put forward in this paper. The metaheuristic AHCOA falls under the category and is known under the broad term of evolutionary algorithms. This is derived by looking at how the ants function and build their nest. Not more on food collection as shown in other terms using nesting in papers. Still the earlier algorithms such as firefly, cuckoo search algorithm, ant colony optimization and the more older and simpler algorithms such as particle swarm optimization and genetic algorithms suffer from some loopholes and downgrading terms of their applicability in diversification(exploration) and exploitation. These are used to create and use them in terms of parent and offspring which are together interspersed together with the fitness value to obtain the output.

Figure IV: Antenna array designer in MATLAB is used in this simulation
The value of dBi versus samples for the linear array design as follows.

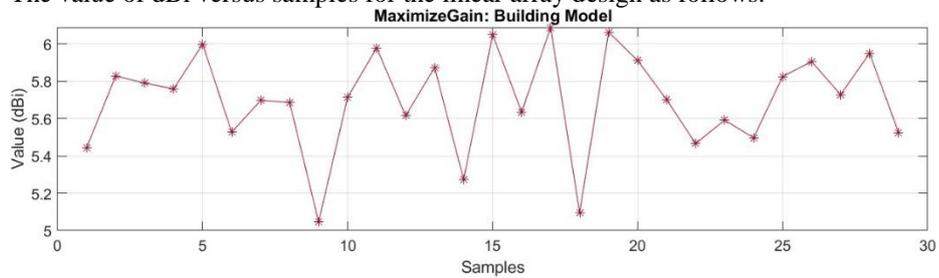

Figure V:Number of elements are considered to be 2

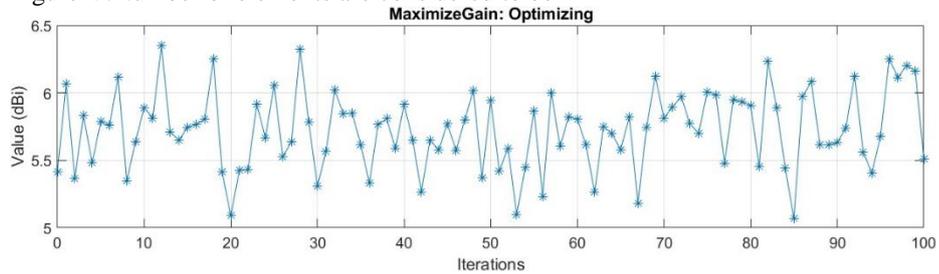

Elements spacing(m) is 0.0045423, lower bound 2 and upperbound 10, AmplitudeTaper(V), lowerbound 1 to 2.4 upperbound, PhaseShift(degrees) lowerbound 90 and upperbound 180.

Construction of the ant hill nest:

Figure VII:

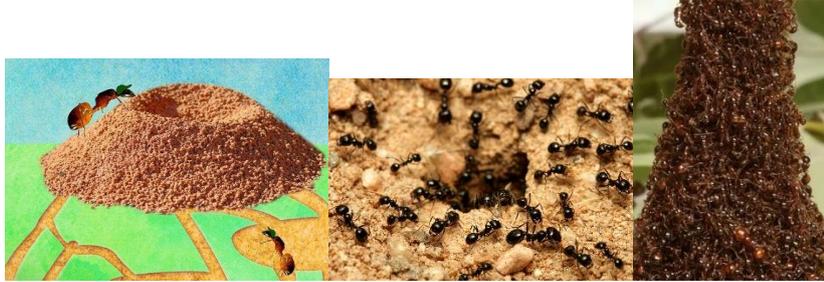

Volume rate pyramid is given by $V = 1/3 * B * h / t$     (2)

Volume rate of right circular cone is given by $V = 1/3 * \pi * r^2 * h / t$     (3)

Figure VII in first image shows about ant hill construction, the second image shows about ants colony before the construction of the ant hill by the species of ants known as Formica ants. The third image showing the process of construction of the ant hill by the entire population of ants. The ants excavate the soil building the ant mold like structure by burying through the soil and then borrowing, eating and throwing the sand in levels.

In this we perform natural selection of the ants for selecting the ants for construction of the ant hill. Randomly guessing is done, as we know that experiments are too much dependent on chance. We select a few population of ants, realize that the ants are similar to certain extent (in construction), start mixing them to get the entire population in building the ant hill. Then we introduce variations in building faster and better ant nest or hills.

*Initialize the first population of Formica ants for ant hill construction*
*Calculate the fitness of ants for the architecture*
*Find the best Formica ants and realize their fitness(determined optimum value)*
*We try to find the global optimum from the natural selection*
***While*** *the construction is not complete and the process is started to satisfy end result*
***For*** *every ant*
*Select a Formica ant using natural selection*
*Update t using equation (2) and (3)*
*Select the shape of the ant bold rectangular or right circular cone*
*Update the population of ants and also*
*Update the position of the ants by random walk by equation (1)*
***End for***
*The ants fitness value for the optimal proximity of the ant hill rate construction*
*Replace the Formica ant with its next ant when the fitness value decreases*
***End while***
***Return*** Rate

*Hypothesis of the AHCOA algorithm*:

Hypothetically the AHCOA is valid in obtaining the global values of the natural selection from the fitness function of the Formica ants.
   a. The ants random walk for collecting food, cleaning, breeding department.
   b. Random walk for construction of the ants hill, while others collect food, build the inner layers handling all activities.
   c. Each population handles each department.
   d. Exploitation of search space is guaranteed by the adaptive construction of the ant hill architecturally.
   e. There is very high chance of stagnation due to the utilization of natural selection.
   f. The AHCOA is a population based algorithm, so the local avoidance of the value is probabilistically quite high.
   g. The ants populations is having a high tide in the beginning and architecture construction rate gradually decreases leading to the convergence of the AHCOA.
   h. Calculating department of the ants for construction determines the diversity of the Formica ants.
   i. AHCOA determines the position of the fittest ants in construction of the ant hill.
   j. In AHCOA the best Formica ants are selected from a search space.
   k. The construction is iterated for different times till the construction gradually completes.

**IV  Experiments and Results:**

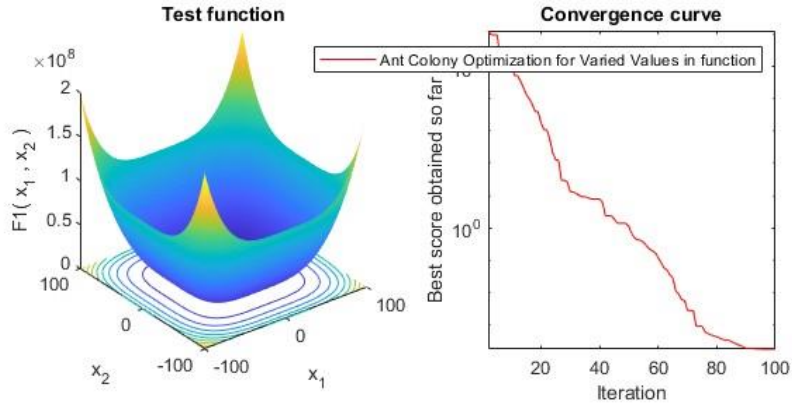

Figure VIII: Function values are completely altered in this case.

| Simulation values for convergence curve | Best solution obtained |
|---|---|
| Iteration 10 | 0.068656 |
| Iteration 20 | -0.061058 |
| Iteration 30 | 0.040118 |
| Iteration 40 | -0.086702 |
| Iteration 50 | 0.079226 |
| Iteration 60 | -0.063675 |
| Iteration 70 | -0.042739 |
| Iteration 80 | -0.027596 |
| Iteration 90 | 0.065217 |

The value at iteration 5 shows a slight variation from the other.

Operators of the AHCOA algorithm:

The ants build the ant hill like an expert architect, there are individual compartments in the ant hill, with each species maintaining each of them. There is some ant species involved in breeding, some in collecting food, some compartment in cleaning the ant hill. There are ant hills build on the ground and there is also ant nest built on the trees which are built by completely different species of ants.

A random walk in modelled as follows when they collect for food and build nest.

$$X(t) = [0, cumulativesum\ 2r(t_1) - 1, cumulativesum\ 2r(t_2) - 1, cumulativesum\ 2r(t_3) - 1, \ldots\ldots cumulativesum\ 2r(t_n) - 1] \quad (1)$$

Random walk is a stochastic function defined as under

Figure IX:Random walk of ants

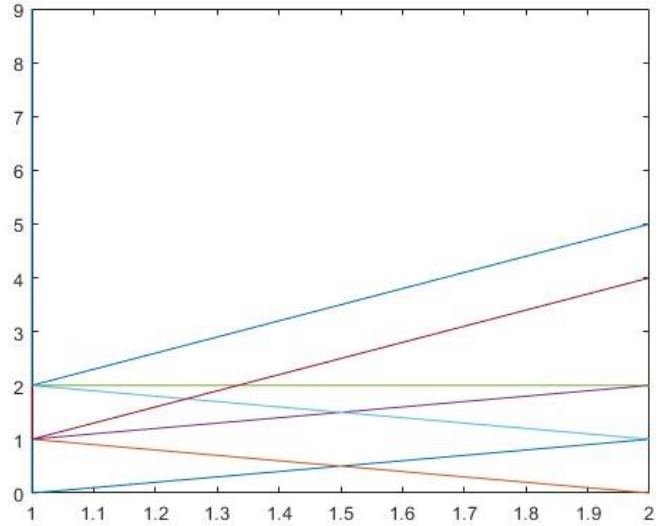

There are different categories of ant colony, which include the breeding queen ant, then the soldier protecting ants, the food collection ants and then the ant hill buildup ant species. These ant species are very distinctly involved in various tasks which are predominantly performed by the ant species. Each one of the ant complements the other and without each others help the colony cannot subsist. When this takes place the ant colony functions very smoothly and perfectly in the system. The random walk of ants was simulated by Mirjalili et.al in his ant colony and ant lion optimizer algorithm. This was inturn further developed by Saxena et.al in optimal antenna synthesis paper, then by Durbadal mandal in his linear array synthesis paper. The antenna system functions very distinctly with the antenna system, in which the sidelobes are tried to be minimized along with controlling the nulls of the antenna system. The side lobes and null depths position are tried to be changes in our paper. The algorithm proves to be very effective in pinear antenna pattern synthesis. In our next paper on NOABS we have distinguishably tried to extract the simulation plots to synthesize the antenna array.

The scatter matrix is given by the following schematix:

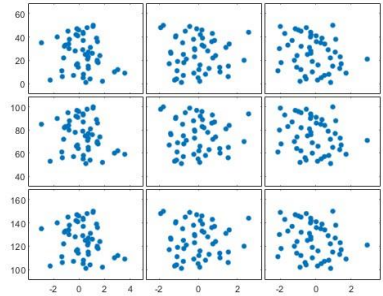

Figure X:

.Figure XI:Results obtained by Saxena et.al in their paper in comparison of uniform array particle swarm optimization in red dotted line

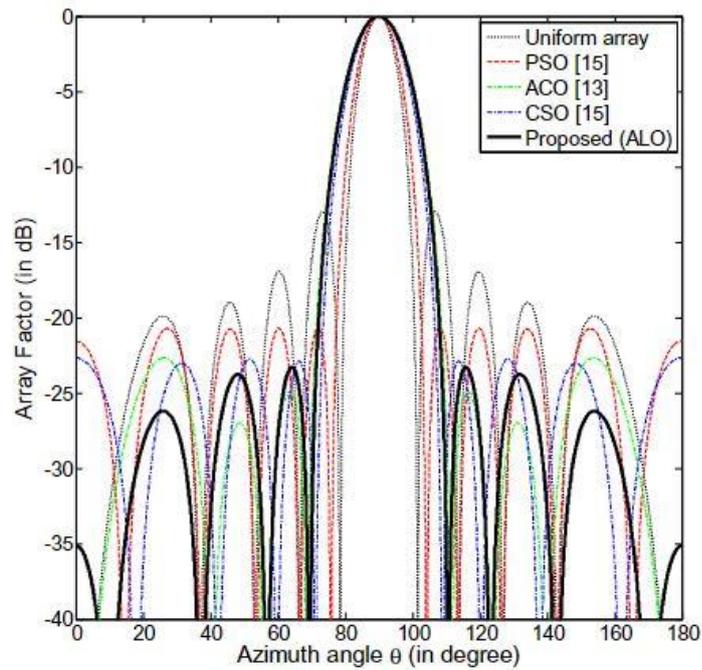

The paper by Saxena et.al shows 8 side lobes while in our paper using AHCOA shows a grating lobe and 8 side lobes and 2 nulls.

Figure XII:Results obtained by AHCOA for 10 element d/λ=0.5

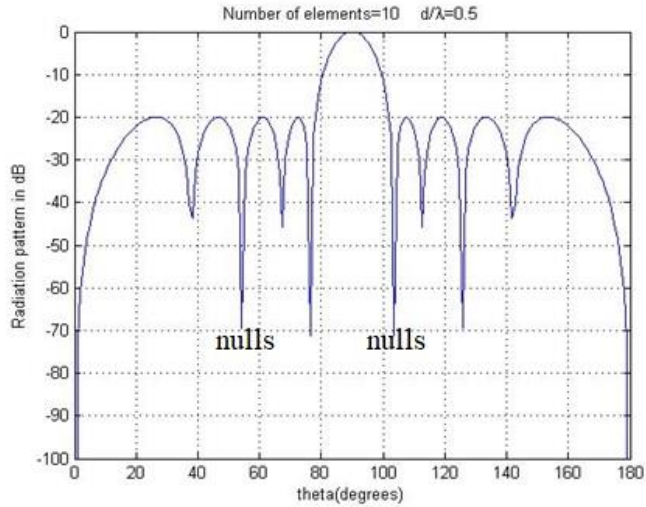

**V Conclusion and results:**
**Figure XIII:** Radiation pattern for AHCOA for inter element gap 0.5.

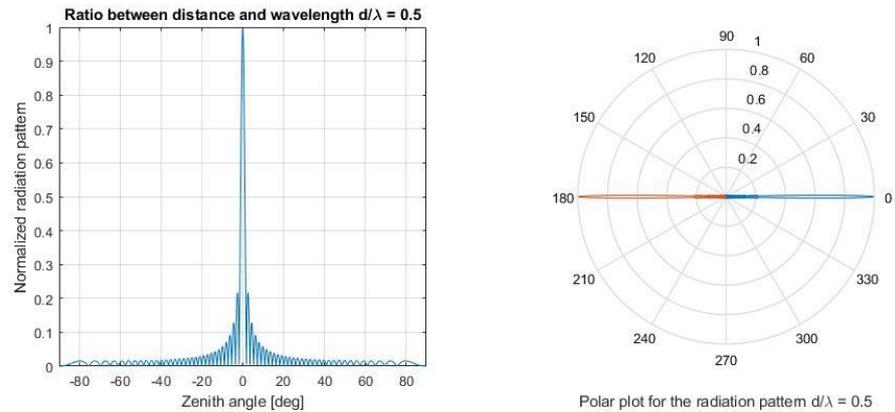

This paper proposes AHCOA. This work and study presented on a novel AHCOA algorithm which is based and inspired by ant hill architecture The proposed method names as AHCOA included initialization, exploration and exploitation to simulate the AHCOA ant hill building algorithm. AHCOA was found to be enough competitive with other state of the art meta-heuristic algorithms. This random precedence of chance is used to figure out which output is the best, this is in turn used to figure out in reducing the side lobes and enhancing the directivity. AHCOA was very competitive with metaheuristic optimizers and superior over conventional techniques. The method mimicked the ant building behavior of the ants. There are complex individual departments in the ant hill involved in food collection, breeding by queen species, cleaning and the Formica species are involved in ant nest or ant hill building construction. First the results on the unimodal functions showed the superior exploitation of the AHCOA algorithm. Second, the exploration ability of AHCOA was confirmed by the further multimodal functions and simulations. Finally in the third observation the results of the composite functions showed high local optima avoidance. Also the convergence function confirmed the convergence of this function in the figure I and VIII but not in figure II. The AHCOA showed high performance in these antenna challenging radiation synthesis.

For future work we are going to exploit more nature based algorithms in antenna synthesis and improve the AHCOA further in solving more engineering problems.